\documentclass{Interspeech}

\interspeechcameraready

\title{Streaming Non-Autoregressive Model for Accent Conversion and Pronunciation Improvement }

\author[affiliation={1}]{Tuan-Nam}{Nguyen}
\author[affiliation={1,2}]{Ngoc-Quan}{Pham}
\author[affiliation={1}]{Şeymanur}{Akti}
\author[affiliation={1,2}]{Alexander}{Waibel}
\affiliation{}{Karlsruhe Institute of Technology}{Germany}
\affiliation{}{Carnegie Mellon University}{USA}
\email{tuan.nguyen@kikt.edu}

\usepackage{comment}
\usepackage{algorithm}
\usepackage{algpseudocode}

\begin{document}

\maketitle


\begin{abstract}
We propose a first streaming accent conversion (AC) model that transforms non-native speech into a native-like accent while preserving speaker identity, prosody and improving pronunciation. Our approach enables stream processing by modifying a previous AC architecture with an Emformer encoder and an optimized inference mechanism. Additionally, we integrate a native text-to-speech (TTS) model to generate ideal ground-truth data for efficient training. Our streaming AC model achieves comparable performance to the top AC models while maintaining stable latency, making it the first AC system capable of streaming.  
\end{abstract}

  %  Effective communication in video conferencing is often hindered by strong accents or mispronunciations. This paper introduces a novel streaming system for accent conversion and pronunciation improvement, designed to enhance intelligibility. The proposed framework leverages a deep learning-based speech synthesis model, capable of dynamically adjusting accent features while preserving the speaker’s natural voice characteristics and prosody. Extensive evaluations on a variety of accents demonstrate significant improvements in speech clarity and comprehensibility, validated by objective metrics and user studies. This system represents a step toward breaking linguistic barriers in digital communication.

\section{Introduction}

Second-language L2 English learners often have accents and mispronunciations that impact communication. AC modifies speech to enhance intelligibility while preserving content, emotion, and speaker identity. Streaming AC enables streaming applications like video conferencing \cite{waibel2012simultaneous}, where seamless interaction are crucial. Previous AC mprocess full utterances, leveraging long context for accurate accent conversion and pronunciation correction.   

Streaming AC requires on-the-fly speech conversion, posing challenges for previous AC models, which process entire utterances at once. Unlike non-streaming models, streaming AC must operate incrementally on speech frames or chunks. In this work, we modify a state-of-the-art non-streaming AC model to support streaming by adjusting its architecture and training process while leveraging knowledge distillation. To our knowledge, this is the first streaming AC model capable of both accent conversion and pronunciation improvement.

Previous AC methods typically follow two approaches. The first involves mapping non-native accents to native accents, relying heavily on parallel data for training \cite{quamer22_interspeech, nguyen24_syndata4genai, 9477581, AC_NamNguyen, jia2024convert}. However, such data—utterances from the same speakers in different accents—is rare and difficult to obtain. To address this, it is possible to use TTS ~\cite{nguyen24_syndata4genai,10096431} or VC \cite{AC_NamNguyen} to generate ground-truth audio synthetically, but these often mismatch in duration and prosody. Seq2seq encoder-decoder \cite{schultz1996lvcsr, huber2022code,schultz2001experiments}models with attention help align inputs and outputs but they still struggle with length mismatches, making them unsuitable for tasks requiring precise synchronization, like video dubbing and conferencing. Their slow autoregressive inference and unstable attention mechanism further limit their applicability to real-time streaming.

The second approach, disentangle-resynthesis, leverages non-parallel data by decomposing speech into components like speaker identity, content, prosody, and accent, then allowing controlled modification and synthesis \cite{jia2023zeroshot,jin2022voicepreserving}. Content features are often represented by bottleneck features (BNFs) extracted from self-supervised models \cite{9814838,NEURIPS2020_92d1e1eb,10.1109/TASLP.2021.3122291}, ASR bottlenecks \cite{jia2023zeroshot}, or ASR logits  \cite{10094737}. However, BNFs inherently capture accent features and pronunciation errors. While adversarial training  \cite{jin2022voicepreserving} and Pseudo-Siamese networks \cite{jia2023zeroshot} attempt to remove accent influences, they struggle to improve pronunciation due to the lack of ground-truth references.

Recent AC advancements combine both disentangling and mapping approaches into a unified framework\cite{nguyen2024improvingpronunciationaccentconversion}. They hypothesize that a TTS system trained solely on native speech will produce accent-independent linguistic representations. Additionally, this native TTS system is able to generate ideal ground-truth data for non-native speakers, ensuring native pronunciation, same speaker identity, duration, prosody, and precise alignment with the original non-native audio. Their AC model leverages TTS text representations to learn accent-independent features and uses synthetic ground-truth to learn the mapping function from non-native accent speech to native-like speech. This approach enables both AC and pronunciation correction while maintaining fast inference due to its non-autoregressive architecture.

Although this model achieves fast inference, it still requires the entire input for prediction, making it unsuitable for streaming applications. We adopt their approach of using native TTS to generate ideal ground truth. Recognizing this model as one of the best for AC, we focus on  architectural and training modifications to develop a streaming-capable AC framework while maintaining the performance of its non-streaming counterpart.

Our work proposes the streaming AC by adapting and enhancing existing non-streaming AC models to operate in a streaming context. Our major contributions include significant architectural and training modifications that enable the model to process speech incrementally while maintaining high performance comparable to non-streaming models. Furthermore, we provide the implementation source code for streaming inference. This marks a substantial step forward in the field, as our streaming AC model is the first to effectively perform AC and pronunciation improvement in streaming fashion. 
\vspace{-0.5cm}
\section{Methodology}
This section presents the step-by-step training process for our streaming AC model. We begin by training the Native TTS model and demonstrating how it generates ideal ground-truth data for non-native speech. Next, we outline the architectural modifications needed for streaming AC and describe the training process using synthetic ground truth. Finally, we detail the inference procedure for streaming.
\subsection{Training Native TTS with prosody and speaker preserving}
\subsubsection{Training}
\begin{figure}[t!]
  \centering
  \includegraphics[width=0.46\textwidth]{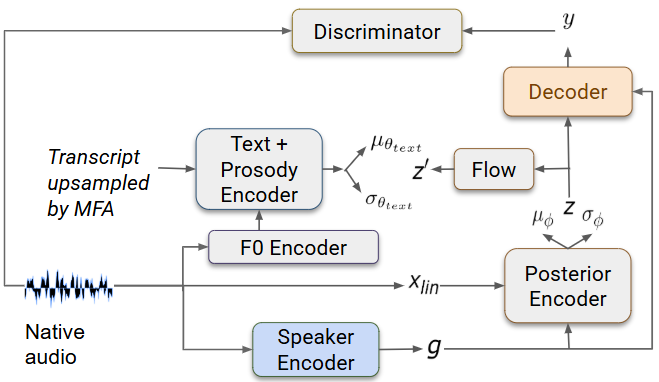}
  \caption{Native TTS with prosody and speaker preserving}
  \label{fig:tts_native}
  \vspace{-0.5cm}
\end{figure}

We train the native TTS model using the VITS \cite{kim2021conditional}, a conditional variational autoencoder (CAVE) enhanced with normalizing flow. VITS consists of three key components: a posterior encoder define \( q_\phi(z|x) \), a prior encoder define  \( p_\theta(z|c) \), and a waveform generator define \( p_\psi(y|z) \),  shown in Figure \ref{fig:tts_native}. 

 %The posterior encoder \( \phi \) models \( q_\phi(z|x) \), while the HiFi-GAN waveform \cite{NEURIPS2020_c5d73680} generator \( \psi \) reconstructs the waveform \( y \) from the latent variable \( z \). The prior encoder \( \theta \) estimates \( p_\theta(z|c) \), with normalizing flow \( f \) refining the latent variables, conditioned on text input \( c \).

To align text and audio, VITS utilizes Monotonic Alignment Search (MAS). To speed up training and eliminate alignment learning, we employ the Montreal Forced Aligner (MFA) \cite{mcauliffe17_interspeech} to upsample text representations before training, ensuring they match the length of the audio. Our prior is conditioned on both the upsampled transcript and the F0 sequence. 

The F0 Encoder extracts frame-level F0 embeddings, which are combined with text embeddings before being processed by a Transformer-based encoder. This Transformer-based encoder \cite{NIPS2017_3f5ee243}, enhanced with normalizing flow, generates the prior distribution \( p_{\theta_{text}}(z|c) \) conditioned on text and prosody information. The posterior encoder uses linear spectrograms \( x_{lin} \) and speaker embeddings \( g \) to sample a latent variable \( q_\phi(z|x) \), while the HiFi-GAN waveform generator \cite{NEURIPS2020_c5d73680} \( \psi \) reconstructs the waveform \( y \) from the latent variable \( z \).

We use a downsampling rate of 320 for computing mel-spectrograms, F0, and linear spectrograms, while the HiFi-GAN decoder applies an upsampling rate of 320 accordingly. F0 sequences are extracted using the YAAPT algorithm \cite{kroon2022comparingconventionalpitchdetection} with the same downsampling rate, and the F0 encoder follows the approach in \cite{wang21n_interspeech}. We utilize the pre-trained speaker encoder from \cite{10.1109/TASLP.2021.3076867}. Our training loss and hyperparameters is similar to the original paper \cite{kim2021conditional}.

\subsubsection{Generate ground-truth from native TTS}

The native TTS model trained earlier is used to generate ideal ground-truth audio for each non-native input. For each non-native audio, we use the MFA to find the alignment and upsample the phoneme transcript match the length of audio. Next, similar to inference process of VITS,  we obtain the latent variable \(z^{'}\) by sampling from  \(p_{\theta_{text}}(z^{'}|c_{upsample})\), which is then refined through an inverted normalizing flow \(f^{-1}\). Finally, the ground-truth audio \(\hat{y}_{\text{ground-truth}}\) is generated from \(p_\psi(x| f^{-1}(z^{'}), g, F0)\) using the HiFi-GAN decoder, where speaker embedding \(g\) and \(F0\) are extracted from original audio. In summary, the synthetic ground truth \({y}_{ground-truth}\) 
is generated from the transcripts with perfect native pronunciation, while utilizing the MFA alignment, \(F0\) and \(g\) from the non-native audio to retain the original duration, prosody and speaker identity.
\begin{equation}
A = MFA (x_{non-native}, c_{text})
\label{eq6}
\end{equation}
\begin{equation}
c_{upsample} = upsampling(c_{text},A)
\label{eq7}
\end{equation}
\begin{equation}
z^{'} \sim p_{\theta_{text}}(z^{'}|c_{upsample}, F_{0})
\label{eq8}
\end{equation}
\begin{equation}
y_{ground-truth} = {Hifigan( f^{-1}(z^{'} ), g, F0)}
\label{eq9}
\end{equation}
\subsection{Streaming non-autoregressive model}
\subsubsection{Architecture}

CVAE architectures are also widely used and proved efficiency in speech-to-speech tasks such as voice conversion (VC)~\cite{10095191} and AC~\cite{nguyen2024improvingpronunciationaccentconversion}.
Unlike TTS, which uses text input for prior encoder, the AC or VC model takes content representation as input. This content representation can be derived from speech recognition models or self-supervised speech models. While the TTS process from text to audio is inherently a one-to-many problem, making the use of a conditional VAE to generate text-conditional prior distributions highly effective, AC or VC involves a one-to-one mapping from rich continuous content representation to audio. As a result, a CVAE may not be strictly necessary for this task. Therefore, to facilitate more straight forward training and inference for streaming AC, we simplified the model by using conventional autoencoder architecture, which include only four main components: a content encoder, a bottleneck extractor, a HiFi-GAN decoder, and a speaker encoder. Our content encoder is designed to extract both content representations and frame-based prosody features, eliminating the need for a separate prosody encoder like in the original non-streaming model. This content encoder is a pre-trained model that remains frozen during the training of the AC model. The bottleneck extractor eliminates accents and corrects pronunciation of content representation, while the HiFi-GAN decoder reconstructs native-like audio from the bottleneck extractor's output while injecting speaker embedding from the original speech.   

The original AC content encoder utilized a Transformer-based ASR model to extract content representations from input audio. However, the conventional Transformer's global attention mechanism is not ideal for streaming applications. To overcome this limitation, we replaced it with Emformer \cite{Hao_EMFformer_MICCAI2024}, a Transformer variant optimized forlow latency  streaming ASR \cite{nguyen21c_interspeech, niehues2018low,niehues2016dynamic}  with lower latency. The Emformer is trained with 12 layers, a hidden size of 1024, a right look-ahead context of 8, a segment size of 4, and a left context of 30.

The bottleneck extractor was originally developed using a WaveNet-based \cite{oord2016wavenetgenerativemodelraw} architecture with non-causal dilated convolutions, which demonstrated strong performance in its initial implementation. Although the bottleneck extractor and original HiFi-GAN decoder are not explicitly designed for streaming, their fully convolutional architecture inherently supports chunk-by-chunk processing, enabling streaming. %While the original HiFi-GAN components are non-causal and not explicitly designed for streaming, their fully convolutional architecture inherently allows for chunk-by-chunk processing, enabling streaming.   

\begin{figure}[t!]
  \centering
  \includegraphics[width=0.40\textwidth]{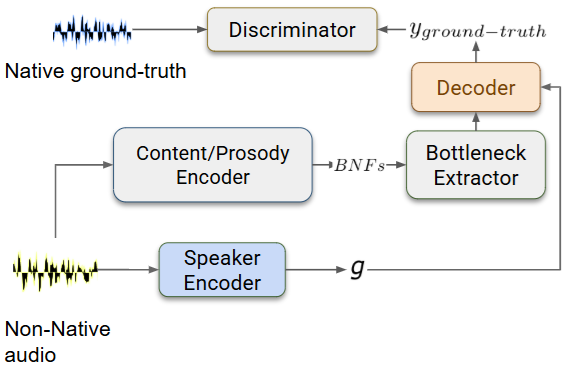}
  \caption{Training streaming Accent Conversion Architecture with generated ideal ground-truth}
  \label{fig:streaming-accent}
    \vspace{-0.7cm}
\end{figure}

To leverage the high-quality audio generation capabilities of previous AC or VC models  in a streaming setup, we retain  both the WaveNet-based bottleneck extractor and HiFi-GAN decoder in our streaming model. To improve inference speed, we replace HiFi-GAN V1 from previous AC model with HiFi-GAN V2. Furthermore, we optimize the source code to facilitate efficient streaming inference for both components.

\subsubsection{Training}
The content encoder is trained on the Librispeech \cite{7178964} and L2Arctic datasets using CMU-dict phoneme labels, optimizing with CTC and prosody losses. This training allows the final layer's output to capture linguistic content and prosody features. The prosody loss is computed as the L1 loss between the predicted and ground truth log-F0 values. Based on our observations, the optimal coefficient for the prosody loss is set to 0.2. The loss function is described below:
\begin{equation}
L_{ASR} = (1- \alpha) * L_{ctc} + \alpha *  L_{prosody}
\end{equation}
We implemented the Emformer based on the Torch Emformer implementation. We observed that the attention padding mask in the Torch Emformer exhibited an average sparsity of 70\%, leading to substantial unnecessary computations in the multi-head scaled dot-product attention operation. To address this, we developed a custom implementation of the Torch Emformer using a Flex Attention function\cite{dong2024flexattentionprogrammingmodel} , significantly optimizing the process. This improvement resulted in a threefold reduction in training time and enabled an eightfold increase in batch size compared to the original implementation. We also provide the source code for this enhanced implementation.

After training the content encoder, we pre-train the streaming AC model exclusively on native data, where both the input and output originate from the same native audio, allowing it to fully learn the distribution and characteristics of native speech. In the second stage, we fine-tune the pre-trained AC model using non-native input paired with the native generated ground-truth output. During this step, both native and non-native data are incorporated into the training process, following a 3:1 ratio between non-native and native audio. For native audio input, the output remains identical to the input, similar to the pre-training phase. This mixed-data approach ensures that the model effectively performs accent conversion and pronunciation correction for non-native speech while retaining the ability to recognize and preserve the accent and pronunciation of native speech, preventing unnecessary modifications. The loss function for both training stages is the HiFi-GAN loss, which consists of mel-spectrogram loss, feature loss, and discriminator loss.

\subsubsection{Streaming Inference}
% All components of the model are non-causal, meaning they require access to future frames to predict the current frame. As we computed, the model requires a look-ahead of 0.64s to predict the current output chunk. During inference, we feed the model input in chunks, each with a duration of 0.08s (corresponding to the size of 4 segment lenght of Emformer model).

It is noticeable that, during training the model has access to the future original frames to predict the current accent-converted frames. Being a fundamentally non-causal model, the look-ahead window is set at 0.64 seconds, accounting for the right context window of the Emformer content model, as well as the Hifi-GAN and WaveNet models. While we have access to the whole utterance during training, in inference the model receives chunk-based inputs, limited to 0.08 seconds (corresponding to the size of 4 segment lengths of the Emformer Model). Our streaming inference strategy is designed to ensure the output remains identical between streaming inference and full-segment inference, minimize the input latency, as well as the smoothness of the streaming output.

In practice, the model inference depends on a Voice Activity Detector (VAD) \cite{sarfjoo21_interspeech} to identify the range of speech within the input stream. Given the detected speech, in order to satisfy the desiderata above, the system waits for a total delay of 0.8 seconds (corresponding to 10 chunks of 0.08s) of audio before outputting the first two converted chunk and extracting the speaker embedding, which remains consistent throughout the entire speech segment. The player thread starts when the model outputs the second chunk, ensuring smooth and uninterrupted audio playback. Details of the streaming inference process are provided in Algorithm \ref{ago:1}

% Before running the model, we employ a Voice Activity Detector (VAD) to identify the start and end times of the speech. When speech{"originatingScript":"m2","payload":{"guid":"9a5f9d28-6c94-4bff-a24e-b5e10ecf90e15b8e9d","muid":"1a8ee153-6a60-4b16-82ea-42e185928d0a78c2dd","sid":"cd415e3c-e6be-423d-904c-c8431fd40805b559aa"}} is detected, the model waits until at least 0.72s (corresponding to 9 chunks of 0.08s) of audio is available before predicting the first two output chunk (around 0.16s) and extracting the speaker embedding, which remains consistent throughout the entire speech segment. Once the first two chunks of streaming output are generated, the audio player's thread begins playing the first output chunk and play the next chunk continuosly.

\begin{algorithm}

\caption{Streaming AC Inference}
\begin{algorithmic}[1]
\label{ago:1}
\State \textbf{Initialize:} output\_chunks $\gets$ [], input\_chunks $\gets$ [], cache $\gets$ None
\While {start stream}
    \State chunk $\gets$ receive\_new\_chunk()
    \State input\_chunks.append(chunk)
    
    \If {len(input\_chunks) $>=$ 10}
        \If {cache == None}
            \State out, cache $\gets$ model.forward(input\_chunks)
        \Else
            \State out, cache $\gets$ model.forward\_with\_cache(chunk, cache)
       \EndIf
        \State output\_chunks.append(out)
    \EndIf

    \If {len(output\_chunks) $==$ 2}
        \State player\_thread.start\_play()
    \EndIf
\EndWhile
\end{algorithmic}
\end{algorithm}

\vspace{-1em}
With a system-wide real-time factor for each chunk below 1 (achievable at  around 0.25 on a GTX 1060 6GB), our streaming process ensures the chunks of output thread remains at least two chunks ahead of the player thread. As a result, the latency stays around 0.8s, regardless of the computational device, enabling smooth and efficient streaming. While latency could be reduced by shortening the look-ahead receptive field, this is beyond our scope. Instead, we focus on an efficient streaming framework that ensures high audio quality.

Notably, streaming inference is performed without any paddings, ensuring that the streaming output remains identical to full-segment processing. After each chunk is processed, the model caches (including Emformer, WaveNet-based component, and HiFi-GAN) are updated and stored, enabling efficient processing of subsequent chunks.

\section{Experiment and Result}
\subsection{Data}

We use the LJSpeech dataset \cite{ljspeech17}, which contains recordings from a single native speaker with consistent pronunciation. To augment the dataset, we employ FreeVC \cite{10095191}, a voice conversion model, to generate multi-speaker native-accented utterances from the original voices. The augmented multi-speaker dataset used in our work consisted of approximately 65,000 utterances from 100 VCTK speakers. These were generated by augmenting the 13,000 utterances from the LJspeech dataset, with each utterance paired with five random speaker embeddings from the VCTK corpus \cite{yamagishi2019vctk}. This ensured a diverse representation of voices with one specific native-like pronunciation. This multi-speaker native dataset is then used to train the native TTS model in a multi-speaker setting and to pre-train the AC model in the first stage.  

In the second stage, we fine-tune the AC model using the L2-ARCTIC dataset \cite{zhao18b_interspeech}, which comprises speech from 24 accented speakers across six different accents. For each accent, we select three speakers for training, while the remaining speakers are used for testing. Each speaker's data is divided into a training set of 1,032 non-overlapping utterances, a validation set of 50 utterances, and a test set of 50 utterances. The test set is selected based on our competitive ASR model~\cite{pham20_interspeech}, focusing on utterances with an average Word Error Rate (WER) greater than 10 across all speakers, under the assumption that higher WERs indicate stronger accents and more pronunciation issues.

\subsection{Evaluation metrics}

\subsubsection{Subjective tests}

\textbf{Nativeness and Speaker Similarity Tests:} Ten participants performed two evaluations using a 5-point scale (1-bad, 2-poor, 3-fair, 4-good, 5-excellent). In the speaker similarity test, they rated how closely the voice identity of the converted audio matched the original input. In the nativeness test, they assessed how native-like the converted speech sounded. 

\subsubsection{Objective tests}

\textbf{Word error rate (WER):} To assess pronunciation improvement, we measure WER using a competitive seq2seq Transformer ASR model \cite{pham20_interspeech}, where lower WER indicates better intelligibility and pronunciation.

\textbf{Accent classifier accuracy (ACC): } We assess accent conversion effectiveness using an accent classifier trained to distinguish native from non-native speech, following the training setup of \cite{AC_NamNguyen}. This metric is measured on both original non-native and converted native-like audio. A larger accuracy gap, where the classifier easily identifies the accent in original audio but struggles with the converted speech, indicates a stronger AC model in reducing non-native accent characteristics.

\textbf{Speaker Embedding Cosine Similarity (SECS): }Speaker similarity is evaluated using SECS, which computes cosine similarity between speaker embeddings of the original and converted speech. We extract embeddings using a state-of-the-art speaker verification model \cite{9814838}, where a similarity score above 0.85 suggests the audios are likely from the same speaker.

\textbf{Mean Opinion Score Net (MOSNet):}  For output quality assessment, we use MOSNet \cite{ROSENBAUM2023159} to predict scores highly correlated with human MOS ratings.

%\textbf{Word error rates (WER), Accent classifier accuracy (ACC) and Speaker Embedding Cosine Similarity (SECS), Mean Opinion Score Net (MOSNet)} To assess pronunciation improvement, we measure Word Error Rate (WER) using a competitive seq2seq Transformer ASR model \cite{pham20_interspeech},  where lower WER indicates better intelligibility. Speaker similarity is evaluated using SECS, which computes cosine similarity between speaker embeddings of the original and converted speech. We extract embeddings using a state-of-the-art speaker verification model \cite{9814838}, where a similarity score above 0.85 suggests the audios are likely from the same speaker. For output quality assessment, we use MOSNet to predict scores highly correlated with human MOS ratings.

%We evaluate accent conversion effectiveness using an accent classifier trained to differentiate between native and non-native speech, following the architecture and training setup of \cite{AC_NamNguyen}. This metric is measured on both the original non-native and converted native audio. A higher accuracy gap between these two sets suggests a more effective AC model, demonstrating successful accent transformation.
\vspace{-0.2cm}

\subsection{Experimental setup}
 Our streaming AC model is based on the  fine-tuned version in the second stage. To assess its performance, we compare it with the non-streaming model introduced in \cite{nguyen2024improvingpronunciationaccentconversion}.

Some causal streaming voice conversion models have been proposed \cite{10446863}. However, in our accent conversion and pronunciation enhancement task, a look-ahead context is essential for accurate pronunciation improvement. To demonstrate this, we also train a causal streaming variant by replacing the non-causal CNN and transposed CNN of the Wavenet-based and Hifigan decoder with their causal counterparts. We analyze the audio quality of synthetic ground truth from the native TTS model. The training hyperparameters follow those of the original HiFi-GAN \cite{NEURIPS2020_c5d73680},  with the system trained for 600,000 steps in the pre-training stage and 200,000 steps in the fine-tuning stage. In the both stages, the mel-spectrogram loss converges to approximately 0.3. Sample evaluation audios are available in \footnote{\url{https://accentconversion.github.io/streaming_demo}}.

\subsection{Result}
The objective metric evaluation in Table \ref{tab:result2} shows that the streaming model performs comparably to the non-streaming model. Surprisingly, MOSNet assigns a higher score to our accent conversion output than to the original audio. This suggests that pre-trained MOSNet, which is primarily trained on non-accented speech, may give lower scores to non-native speech, even when it is real. A MOSNet score of approximately 4.1 indicates that our generated audio maintains good quality in terms of naturalness. Additionally, SECS remains stable at 0.85 and 0.84, verifying that the speaker identity is well-preserved in both settings. The ACC results are weaker compared to other metrics. Since our model aims to retain the original audio's prosody, which can sometimes reflect the speaker's accent, it occasionally classifies the converted audio as non-native. This prosody preservation may, in some cases, reduce the effectiveness of accent conversion. The causal streaming model performs worst across all metrics, highlighting the importance of look-ahead context. Subjective evaluation (Table \ref{tab:result1}) shows similar scores for streaming and non-streaming models, while strong WER performance indicates high-quality synthetic ground truth.

\begin{table}[ht]

	\setlength{\tabcolsep}{4pt}
	\centering
 \caption{Subjective metrics}
	\begin{tabular}{lccc}
	 \cline{1-3}
       \textbf{Models} &  {Nativeness} & {Sim-MOS} \\ 
       
        \cline{1-3}
        Original  & \(1.12\pm0.15\) & - \\      
        Synthetic ground-truth & \(3.93\pm0.15\) & \(4.02 \pm0.17\)  \\
         \cline{1-3}
         Causal streaming Model & \(1.23\pm0.12\)  & \(3.75\pm0.12\) \\
         Streaming model & \(3.78\pm0.18\) & \(3.92\pm0.18\)\\
         Non-Streaming model & \(3.87\pm0.20\)  & \(3.96\pm0.19\)\\
         \cline{1-3}
	\end{tabular}

\label{tab:result1}

\end{table}
\vspace{-0.5cm}
\begin{table}[ht]
        \centering
	\setlength{\tabcolsep}{4pt}
	
\caption{Objective metrics}

	\begin{tabular}{lccccc}
	 \cline{1-5}
       \textbf{Models} &  {WER} & {ACC} & {SECS} & {MOSNet}\\ 
       
        \cline{1-5}
        Original  & 18.3 & 98.3 & 1.0 & 3.97\\
      
        Synthetic ground-truth & 4.7 & 11.5 & 0.83 & 4.18 \\
         \cline{1-5}
        Causal streaming Model & 33.5 & 25.1 & 0.75 & 3.01 \\
         Streaming model  & 14.1 & 16.3 & 0.85 &  4.12\\
         Non Streaming model & 14.3  & 17.0 &  0.84 & 4.08\\
       
         \cline{1-5}
	\end{tabular}

\label{tab:result2}

\end{table}

\section{Conclusion}
This work presents the first streaming accent conversion model, demonstrating its capability to synthesize high-quality audio with native-like pronunciation. Our results show that the streaming model achieves performance comparable to its non-streaming counterpart. We introduced an effective streaming inference method for the non-autoregressive accent conversion model. Although the current latency is not yet at a low-latency level (0.8s), it can be further reduced by decreasing the look-ahead receptive field of the entire model. Future work will focus on minimizing look-ahead latency while maintaining audio quality, as well as optimizing inference for low-power computing devices such as CPUs.
\section{Acknowledgment}
This research was supported by a grant from Zoom Video Communications , Inc. T;  European Commission Project Meetween (101135798), the Federal Ministry of Education and Research (BMBF)
 of Germany under the number 01EF1803B (RELATER), and the pilot program Core-Informatics of the Helmholtz A and the HoreKa supercomputer funded by the Ministry of Science, Research and the Arts Baden-Württemberg and BMBF.
\bibliographystyle{IEEEtran}
\bibliography{mybib}

\end{document}